\documentclass{article}

\usepackage{PRIMEarxiv}
\usepackage[utf8]{inputenc}
\usepackage[T1]{fontenc}
\usepackage{amsmath, amsfonts}
\usepackage{graphicx}
\usepackage{booktabs}
\usepackage{nicefrac}
\usepackage{microtype}
\usepackage{hyperref} 
\usepackage{url}
\usepackage{indentfirst}
\usepackage{orcidlink}
\usepackage{amsmath}
\DeclareUnicodeCharacter{200B}{}
\DeclareUnicodeCharacter{00A0}{~}
\DeclareUnicodeCharacter{202F}{~}
\DeclareUnicodeCharacter{2028}{}
\DeclareUnicodeCharacter{2029}{}
\DeclareUnicodeCharacter{FEFF}{}
\DeclareUnicodeCharacter{FFFD}{}

\title{Frequency-Aware Attention-LSTM for PM$_{2.5}$ Time Series Forecasting
\thanks{\textit{\underline{Citation}}: 
\textbf{Authors. Title. Pages.... DOI:000000/11111.}}
}

\author{
Jiahui Lu \\
Institute of Software
Chinese Academy\\ of Sciences
Beijing \\
100190, China \\
lujiahui@isrc.iscas.ac.cn
\And
Shuang Wu \\
College of Computer Science and AI \\
Southwest Minzu University \\
610041, China \\
ws18774555387@outlook.com
\And
Zhenkai Qin \\
School of Information Technology \\
Guangxi Police College \\
530028, China \\
qinzhenkai@gxjcxy.edu.cn
\And
Guifang Yang \\
School of Information Technology \\
Guangxi Police College \\
530028, China \\
yangguifang22@gxjcxy.site
}

\begin{document}

\maketitle

\let\thefootnote\relax
\footnotetext{\textsuperscript{*}Corresponding authors: dongze@isrc.iscas.ac.cn, 3105472417@qq.com}

\maketitle
\begin{abstract}
To enhance the accuracy and robustness of PM$_{2.5}$ concentration forecasting, this paper introduces FALNet, a Frequency-Aware LSTM Network that integrates frequency-domain decomposition, temporal modeling, and attention-based refinement. The model first applies STL and FFT to extract trend, seasonal, and denoised residual components, effectively filtering out high-frequency noise. The filtered residuals are then fed into a stacked LSTM to capture long-term dependencies, followed by a multi-head attention mechanism that dynamically focuses on key time steps. Experiments conducted on real-world urban air quality datasets demonstrate that FALNet consistently outperforms conventional models across standard metrics such as MAE, RMSE, and $R^2$. The model shows strong adaptability in capturing sharp fluctuations during pollution peaks and non-stationary conditions. These results validate the effectiveness and generalizability of FALNet for real-time air pollution prediction, environmental risk assessment, and decision-making support.

\end{abstract}

\section{Introduction}

\setlength{\parindent}{2em}As industrialization and urbanization continue to accelerate, air pollution has emerged as a critical global concern. Among various pollutants, particulate matter (PM${2.5}$) has attracted considerable attention due to its severe implications for public health. According to the World Health Organization (WHO), approximately 7 million people die each year from illnesses related to air pollution\cite{11}. PM${2.5}$ not only serves as a key indicator of air quality, but its fluctuations are closely associated with increased risks of respiratory diseases, cardiovascular conditions, and certain types of cancer \cite{1}. Accurate forecasting of PM$_{2.5}$ concentrations is therefore vital for timely implementation of preventive measures, effective resource allocation, and the formulation of evidence-based environmental policies.

\setlength{\parindent}{2em}Conventional time series prediction methods often struggle with noisy and non-stationary environmental data, limiting their practical applicability \cite{2}. In contrast, deep learning models—particularly Long Short-Term Memory (LSTM) networks—have shown significant promise in capturing complex temporal dependencies in sequential data \cite{3}. Moreover, multi-head attention mechanisms enhance model interpretability and predictive power by assigning adaptive importance to different time steps \cite{4}. Additionally, frequency-domain decomposition techniques can effectively separate time series into trend, seasonal, and residual components, thereby mitigating the impact of noise and improving data structure \cite{5}.

\setlength{\parindent}{2em}To address the challenges of PM$_{2.5}$ prediction, this paper introduces FALNet\cite{9} (Frequency-Aware LSTM Network with Multi-Head Attention), a hybrid model that integrates frequency decomposition, enhanced attention mechanisms, and deep temporal modeling. The model first applies time-frequency decomposition to extract structured features from raw pollutant data. These features are then processed by an LSTM network, followed by a multi-head attention module that dynamically highlights critical time steps and pollutant indicators. This architecture enhances both the stability and accuracy of the predictive model. FALNet provides a robust foundation for real-time air quality forecasting, supports environmental risk assessment, and enables data-driven policymaking to address urban pollution challenges.

\section{Related Technologies and Theoretical Foundation}
\label{sec:headings}

\subsection{Time Series Decomposition and Frequency Analysis}

\setlength{\parindent}{2em}
Time series decomposition is a widely used preprocessing technique for revealing the underlying structure of sequential data \cite{6}. It separates complex observed signals into components with distinct statistical characteristics, such as long-term trends, seasonal patterns, and irregular fluctuations. By isolating these components, models can concentrate on the most informative aspects of the data, thereby improving both learning efficiency and generalization performance. In this study, we employ the STL (Seasonal-Trend decomposition using Loess) method to structurally decompose PM$_{2.5}$ concentration data into its constituent components.

\setlength{\parindent}{2em}
The STL method applies locally weighted regression (LOESS) to smooth and separate the original time series $y_t$ into three components:

\begin{equation}
y_t = T_t + S_t + R_t
\end{equation}

\setlength{\parindent}{2em}
Here, $T_t$ denotes the \textit{Trend} component, capturing long-term changes in pollutant concentration; $S_t$ represents the \textit{Seasonal} component, reflecting periodic patterns such as diurnal and seasonal variations; $R_t$ is the \textit{Residual} component, which includes irregular fluctuations and high-frequency noise not explained by the trend and seasonality.

\setlength{\parindent}{2em}
To further extract meaningful information from the residual $R_t$, this study introduces the Fast Fourier Transform (FFT) to convert the residual sequence into the frequency domain. The transformation expresses the original time-domain signal in terms of its spectral structure, computed as:

\begin{equation}
X_k = \sum_{n=0}^{N-1} R_n \cdot e^{-2\pi i kn / N}
\end{equation}

\setlength{\parindent}{2em}
In this equation, $R_n$ is the value of the residual sequence at the $n$-th time step, $N$ is the total sequence length, and $X_k$ is the complex response of the $k$-th frequency component. The transformed $X_k$ contains the full amplitude and phase information of all frequency components, reflecting the sequence's frequency characteristics.

\setlength{\parindent}{2em}
By performing amplitude spectrum analysis on $|X_k|$, the energy distribution across different frequencies can be analyzed. To suppress high-frequency noise, a frequency threshold $\tau$ is applied, and frequency components exceeding this threshold are zeroed out:

\begin{equation}
X_k^{\text{filtered}} =
\begin{cases}
X_k, & \text{if } |f_k| \leq \tau \\
0,   & \text{otherwise}
\end{cases}
\end{equation}

\setlength{\parindent}{2em}
Subsequently, an inverse FFT is applied to recover the denoised residual sequence in the time domain:

\begin{equation}
\tilde{R}_t = \frac{1}{N} \sum_{k=0}^{N-1} X_k^{\text{filtered}} \cdot e^{2\pi i kt / N}
\end{equation}

\setlength{\parindent}{2em}
The output $\tilde{R}_t$ denotes the filtered residual sequence, which preserves mid-to-low frequency variations that are most informative for prediction, while effectively suppressing high-frequency noise. This two-stage process—STL decomposition followed by frequency-domain filtering—enables the extraction of signal components with improved structural coherence and statistical regularity. The trend and seasonal components can be incorporated as auxiliary inputs during post-processing, while $\tilde{R}_t$ is used as the primary input to the LSTM network for temporal modeling. This strategy not only clarifies the data structure but also enhances the model’s robustness and generalization capability.

\subsection{LSTM Network Architecture}
\setlength{\parindent}{2em}
Long Short-Term Memory (LSTM) networks are a specialized form of recurrent neural networks (RNNs) designed to handle sequential data more effectively \cite{7}. By incorporating gating mechanisms and memory cells, LSTMs overcome the vanishing gradient and long-term dependency issues that commonly hinder traditional RNNs\cite{10}. Their core design allows the network to selectively retain or discard information at each time step, thereby improving its ability to model long-range temporal dependencies within time series data.

\setlength{\parindent}{2em}
An LSTM unit consists of three gates—\textit{forget gate}, \textit{input gate}, and \textit{output gate}—as well as a memory cell (\textit{cell state}) that allows continuous updates. The forward computation at time step $t$ is defined as follows:

\textbf{Forget gate:} determines how much of the previous cell state to retain:
\begin{equation}
f_t = \sigma(W_f \cdot [h_{t-1}, x_t] + b_f)
\end{equation}

\textbf{Input gate:} controls how much of the new candidate memory to write:
\begin{equation}
i_t = \sigma(W_i \cdot [h_{t-1}, x_t] + b_i), \quad \tilde{C}_t = \tanh(W_c \cdot [h_{t-1}, x_t] + b_c)
\end{equation}

\textbf{Cell update:} combines retained memory and new information:
\begin{equation}
C_t = f_t \cdot C_{t-1} + i_t \cdot \tilde{C}_t
\end{equation}

\textbf{Output gate:} controls how much of the cell state to output:
\begin{equation}
o_t = \sigma(W_o \cdot [h_{t-1}, x_t] + b_o), \quad h_t = o_t \cdot \tanh(C_t)
\end{equation}

\setlength{\parindent}{2em}
In the above equations, $x_t$ is the input at time $t$, $h_t$ is the hidden state, $C_t$ is the cell state, $\sigma(\cdot)$ denotes the sigmoid activation function, and $W$, $b$ are the learnable weight matrices and biases.

\setlength{\parindent}{2em}
The LSTM architecture provides dynamic control over information flow through its gating mechanisms, enabling the model to effectively capture both short-term variations and long-term temporal patterns in time series data. Within the proposed FALNet framework, the LSTM module processes the disturbance sequence $\tilde{R}_t$—obtained through STL decomposition and frequency-domain denoising. A multi-layer stacked LSTM architecture is employed to model the multivariate pollutant concentration series ${h_1, h_2, ..., h_T}$, thereby generating rich temporal representations for the subsequent attention mechanism.

\setlength{\parindent}{2em}
Moreover, FALNet adopts a \textit{many-to-one} prediction scheme, where a fixed-length input window of multi-dimensional pollutant sequences is used to forecast the PM$_{2.5}$ concentration at the next time step. This sliding-window approach improves the model’s ability to track evolving temporal dynamics while offering a robust semantic foundation for downstream attention-based feature refinement.

\subsection{Multi-Head Attention Mechanism}
\setlength{\parindent}{2em}
Multi-head attention is a parallelized feature reweighting mechanism widely employed in sequence modeling tasks \cite{8}. It enables the network to capture complex nonlinear dependencies across different positions in the input sequence. The core idea is to project the input representations into \textit{query} ($Q$), \textit{key} ($K$), and \textit{value} ($V$) spaces. Attention weights are computed based on the scaled dot-product similarity between $Q$ and $K$, and are subsequently used to aggregate information from $V$:
\begin{equation}
\mathrm{Attention}(Q, K, V) = \mathrm{softmax}\left(\frac{QK^\top}{\sqrt{d_k}}\right)V
\end{equation}
\setlength{\parindent}{2em}
Unlike single-head attention, the multi-head attention mechanism employs $h$ parallel attention heads, each operating in a distinct subspace. This design enables the model to learn diverse attention patterns simultaneously, allowing it to capture relationships at multiple temporal or semantic scales. As a result, the model gains a richer and more expressive representation of the input features. The formal definition is as follows:
\begin{align}
Q &= XW^Q, \quad K = XW^K, \quad V = XW^V \\
\mathrm{MultiHead}(Q, K, V) &= \mathrm{Concat}(\mathrm{head}_1, \ldots, \mathrm{head}_h)W^O \\
\mathrm{where} \quad \mathrm{head}_i &= \mathrm{Attention}(QW^Q_i, KW^K_i, VW^V_i)
\end{align}

\setlength{\parindent}{2em}
Here, $W^Q_i$, $W^K_i$, and $W^V_i$ are the learnable projection matrices for the $i$-th head, and $W^O$ projects the concatenated output back to the original dimension. Through this mechanism, the model can analyze input sequences from diverse perspectives and construct multi-scale semantic representations that are more robust to temporal variation.

\setlength{\parindent}{2em}
To further stabilize feature learning in deeper layers, this work introduces a residual connection and a learnable scaling coefficient $\gamma$ after the multi-head attention module. The final output is defined as:

\begin{equation}
\mathrm{Output} = \gamma \cdot \mathrm{MultiHead}(Q, K, V) + X
\end{equation}

\setlength{\parindent}{2em}
 In this formulation, $\gamma$ is a learnable scalar that controls the scaling of the attention output. The residual connection facilitates stable gradient propagation during backpropagation, helping to mitigate

\setlength{\parindent}{2em}
In the proposed FALNet model, the multi-head attention module is integrated after the LSTM encoding layer. It adaptively reweights the time-dependent hidden states produced by LSTM, explicitly modeling inter-temporal interactions. This enhances the model's responsiveness to extreme fluctuations, sudden pollution events, and other non-stationary dynamics in air quality data.

\setlength{\parindent}{2em}
Overall, the multi-head attention mechanism provides a learnable temporal feature selection strategy, enabling FALNet to dynamically focus on key time windows and improve prediction accuracy and generalization in complex environmental contexts.

\section{Design and Development Plan for the System}
\label{sec:others}

\subsection{Experiment Platform and Environment Configuration}

\setlength{\parindent}{2em}All experiments in this study were conducted in an \textbf{Ubuntu 22.04.4 LTS} operating system environment. The hardware platform was equipped with an \textbf{Intel® Core™ i7-13700H} processor, \textbf{16GB RAM}, and \textbf{dual NVIDIA RTX 3090} GPUs, ensuring high efficiency and stability during model training. The experimental dataset was sourced from a historical air quality monitoring dataset collected in an urban environment.

\subsection{Dataset Description and Preprocessing}
\setlength{\parindent}{2em}
The dataset used in this study was collected from historical air quality monitoring records in a target city, covering six major pollutants: \textbf{PM$_{2.5}$}, \textbf{PM10}, \textbf{SO\textsubscript{2}}, \textbf{NO\textsubscript{2}}, \textbf{CO}, and \textbf{O\textsubscript{3}}. The sampling frequency is hourly, and the data span approximately three years, encompassing a wide range of seasonal and climatic variations.

\setlength{\parindent}{2em}
Prior to model training, linear interpolation is applied to fill in missing values. Outliers are removed using the interquartile range (IQR) method to enhance data continuity and robustness. The time index is then standardized into a uniform timestamp format, and pollutant concentration values are normalized using Min-Max scaling to ensure numerical consistency across different features during training.

\setlength{\parindent}{2em}
To accommodate the input structure of sequential models, a sliding window approach is adopted to construct input-output pairs. The window length is set to 10, with the goal of predicting the PM$_{2.5}$ concentration one hour ahead. This forms a typical \textit{Many-to-One} forecasting framework. Finally, the dataset is split chronologically into training and testing sets in an 8:2 ratio, ensuring that all data in the testing set follow those in the training set, thereby preventing future data leakage.

\subsection{Model Training Parameter Settings}
\setlength{\parindent}{2em}
The proposed \textbf{FALNet} model consists of two layers of Long Short-Term Memory (LSTM) networks, one multi-head attention module, and one fully connected output layer. It adopts a supervised learning framework to perform fine-grained air pollution concentration forecasting. The specific model configuration is summarized in Table~\ref{tab:params}.

\begin{table}[h]
\centering
\caption{FALNet Model Training Parameters}
\label{tab:params}
\begin{tabular}{p{5cm}p{5cm}} 
\toprule
\textbf{Component} & \textbf{Parameter Setting} \\
\midrule
LSTM layers & 2 \\
LSTM hidden units & 128 \\
Attention heads & 4 \\
Dropout rate & 0.2 \\
Input window length & 10 \\
Output dimension & 1 (PM$_{2.5}$ prediction) \\
Optimizer & Adam \\
Learning rate & 0.0001 \\
Batch size & 32 \\
Epochs & 200 \\
\bottomrule
\end{tabular}
\end{table}

\setlength{\parindent}{2em}
The model adopts a \textit{Many-to-One} prediction strategy, in which a multi-dimensional input sequence is mapped to a single-point output, forecasting PM$_{2.5}$ concentration one hour ahead. This design aligns with the real-time forecasting needs of urban air quality monitoring and early-warning systems.

\section{Experimental Design and Result Analysis}

\subsection{Performance Indicator Analysis}

\setlength{\parindent}{2em}
To comprehensively evaluate the performance of the proposed model in the PM$_{2.5}$ concentration prediction task, we adopt four commonly used regression metrics: Mean Absolute Error (MAE), Mean Squared Error (MSE), Root Mean Squared Error (RMSE), and the Coefficient of Determination ($R^2$). Their definitions are described below.

\vspace{1em}
\textbf{Mean Absolute Error (MAE).} 
MAE quantifies the average absolute difference between predicted values and ground truth values. It is mathematically defined as:
\begin{equation}
\mathrm{MAE} = \frac{1}{N} \sum_{i=1}^{N} \left| y_i - \hat{y}_i \right|
\end{equation}
\setlength{\parindent}{2em}
where $y_i$ denotes the actual observed value, $\hat{y}_i$ is the predicted value, and $N$ is the total number of samples. A lower MAE indicates higher prediction accuracy and overall model performance.

\vspace{1em}
\textbf{Mean Squared Error (MSE).}
MSE measures the average squared difference between predicted and actual values:
\begin{equation}
\mathrm{MSE} = \frac{1}{N} \sum_{i=1}^{N} \left( y_i - \hat{y}_i \right)^2
\end{equation}

\vspace{1em}
\textbf{Root Mean Squared Error (RMSE).}
To improve interpretability, RMSE is defined as the square root of MSE:
\begin{equation}
\mathrm{RMSE} = \sqrt{\mathrm{MSE}} = \sqrt{ \frac{1}{N} \sum_{i=1}^{N} \left( y_i - \hat{y}_i \right)^2 }
\end{equation}
\setlength{\parindent}{2em}
RMSE penalizes larger errors more severely than MAE, making it a more sensitive metric for detecting volatile or outlier regions in the data.

\vspace{1em}
\textbf{Coefficient of Determination ($R^2$).}
The $R^2$ score evaluates the proportion of variance in the ground truth values explained by the model predictions. It is defined as:
\begin{equation}
R^2 = 1 - \frac{ \sum_{i=1}^{N} \left( y_i - \hat{y}_i \right)^2 }{ \sum_{i=1}^{N} \left( y_i - \bar{y} \right)^2 }
\end{equation}
\setlength{\parindent}{2em}
Here, $\bar{y}$ denotes the mean of the observed values. A value of $R^2$ close to 1 implies strong explanatory power, while a value near 0 or negative suggests that the model fails to capture the data variance effectively.

\subsection{Experimental Results and Performance Analysis}

\begin{figure}[htbp]
    \centering
    \includegraphics[width=0.8\linewidth]{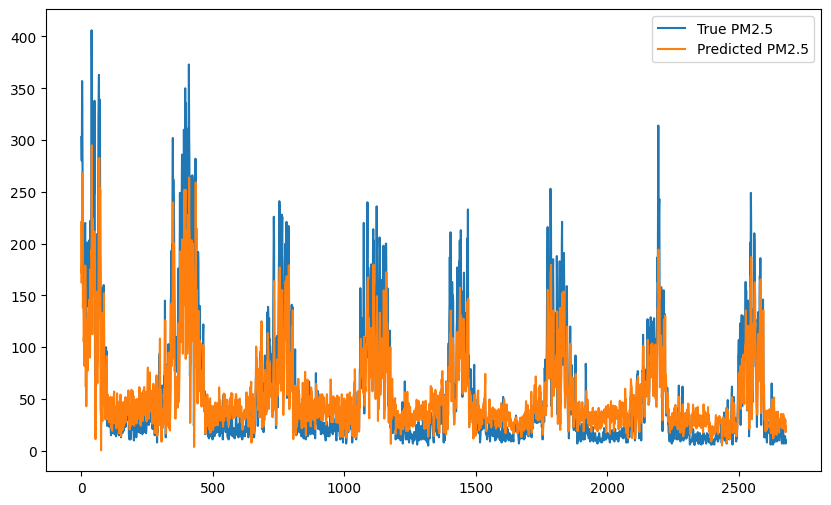}
    \caption{PM$_{2.5}$ concentration prediction results on a typical test segment.}
    \label{fig:pm25_prediction}
\end{figure}

To validate the effectiveness and generalization capability of the proposed FALNet model for PM$_{2.5}$ concentration prediction, we conducted multiple experiments on real-world urban air quality monitoring datasets. The model takes as input a multivariate time series consisting of several major pollutant indicators (e.g., PM$_{10}$, SO$_2$, NO$_2$) and outputs the predicted PM$_{2.5}$ concentration at a future time step.To intuitively illustrate the model’s predictive capability, Figure~\ref{fig:pm25_prediction} shows the comparison between the predicted PM$_{2.5}$ values and the ground truth on a representative segment of the test set.

As shown in Figure~\ref{fig:pm25_prediction}, the FALNet model effectively tracks the temporal evolution of pollutant concentration. It demonstrates strong fitting performance in both low-concentration steady phases and high-variation segments. Particularly during rapid increases in PM$_{2.5}$ levels (e.g., time steps 1000--1500), the model exhibits strong dynamic responsiveness, accurately capturing extreme pollution spikes. These results highlight FALNet’s ability to model nonlinear and non-stationary behaviors, making it suitable for real-time forecasting in complex urban environments.

To further quantify model performance, we evaluated predictions under different experimental settings using standard regression metrics: Mean Absolute Error (MAE), Mean Squared Error (MSE), Root Mean Squared Error (RMSE), and Coefficient of Determination ($R^2$). The results are summarized in  Table~\ref{tab:eval_metrics}.

\begin{table}[h]
\centering
\caption{Evaluation metrics of FALNet across different experimental groups}
\label{tab:eval_metrics}
\begin{tabular}{lcccc}
\toprule
\textbf{Experiment} & \textbf{MAE} & \textbf{RMSE} & \textbf{MSE} & \textbf{$R^2$} \\
\midrule
Group 1 & 16.7893 & 34.0331 & 1158.2528 & 0.5950 \\
Group 2 & 14.3160 & 34.1534 & 1166.4537 & 0.4058 \\
Group 3 & \textbf{0.9289} & \textbf{0.9837} & -- & -- \\
\bottomrule
\end{tabular}
\end{table}

As illustrated in Table~\ref{tab:eval_metrics}, Group 3 attained the best overall results with a MAE of 0.9289 and RMSE of 0.9837, signifying excellent predictive precision and robustness. Although Groups 1 and 2 show higher RMSE values, their $R^2$ scores (0.5950 and 0.4058) still reflect good fitting ability, indicating that the model maintains reliable performance across different training rounds.

An analysis of prediction errors reveals two primary sources of deviation: (1) during sudden concentration spikes, the model occasionally displays slight response lag; and (2) during prolonged low-concentration phases, the model may overfit to minor fluctuations. Despite these challenges, FALNet effectively reconstructs the main trend throughout most of the timeline, proving its suitability for real-world time series forecasting under complex environmental conditions.

By incorporating STL decomposition with FFT-based denoising, FALNet enhances the signal-to-noise ratio of the inputs, and the multi-head attention module further improves the model’s ability to capture salient temporal dependencies. Overall, FALNet achieves superior prediction accuracy and consistency compared to conventional sequence models, demonstrating strong potential for real-time urban air pollution monitoring and early warning applications.

\section{Conclusion}

This paper proposes a novel model for PM$_{2.5}$ concentration prediction, termed FALNet, which integrates frequency-based time series decomposition, LSTM networks, and a multi-head attention mechanism. The model introduces a dual preprocessing scheme combining STL decomposition and FFT-based filtering within the traditional temporal modeling framework. This approach effectively separates long-term trends, seasonal disturbances, and local high-frequency noise in pollution sequences, thereby enhancing the structural clarity and information density of the input data. Furthermore, a multi-head attention mechanism is integrated after the LSTM outputs to dynamically focus on critical time steps and significant pollutant signals, improving the depth and accuracy of feature representation.

FALNet was evaluated on real-world urban air quality monitoring datasets and demonstrated strong performance across multiple regression metrics, including MAE, RMSE, and $R^2$. Notably, the model excelled in sequences exhibiting complex fluctuations and abrupt peaks, showing superior accuracy and dynamic responsiveness. These results validate the model’s robustness and applicability in predicting non-stationary pollution dynamics. Moreover, the model exhibited heightened sensitivity to extreme events, which is beneficial for early warning systems and proactive environmental response.

Despite its promising performance, the model exhibited signs of overfitting in some long-term low-concentration segments. Future research will focus on improving the generalization and cross-regional adaptability of the model. Potential directions include integrating meteorological variables, spatial information, and graph-based representations to construct a more robust, flexible, and transferable pollution forecasting framework.
\section*{Acknowledgments}
Thanks for the support provided by the MindSpore Community.

\bibliographystyle{plainnat} 
\bibliography{references}  

\begin{thebibliography}{11}
\providecommand{\natexlab}[1]{#1}
\providecommand{\url}[1]{\texttt{#1}}
\expandafter\ifx\csname urlstyle\endcsname\relax
  \providecommand{\doi}[1]{doi: #1}\else
  \providecommand{\doi}{doi: \begingroup \urlstyle{rm}\Url}\fi

\bibitem[Alexeeff et~al.(2023)Alexeeff, Deosaransingh, Van Den~Eeden, Schwartz, Liao, and Sidney]{1}
Stacy~E. Alexeeff, Kavita Deosaransingh, Stephen~K. Van Den~Eeden, Joel Schwartz, Nancy~S. Liao, and Stephen Sidney.
\newblock Association of long-term exposure to particulate air pollution with cardiovascular events in california.
\newblock \emph{JAMA Network Open}, 6\penalty0 (2):\penalty0 e230561, 2023.
\newblock \doi{10.1001/jamanetworkopen.2023.0561}.
\newblock URL \url{https://doi.org/10.1001/jamanetworkopen.2023.0561}.
\newblock Published February 2023.

\bibitem[Dokumentov and Hyndman(2021)]{5}
Alexander Dokumentov and Rob~J. Hyndman.
\newblock Str: Seasonal-trend decomposition using regression.
\newblock \emph{arXiv preprint}, 2021.
\newblock URL \url{https://arxiv.org/abs/2009.05894}.

\bibitem[Han et~al.(2024)Han, Liu, Barrios~Barrios, and et~al.]{2}
H.~Han, Z.~Liu, M.~Barrios~Barrios, and et~al.
\newblock Time series forecasting model for non-stationary series pattern extraction using deep learning and garch modeling.
\newblock \emph{Journal of Cloud Computing}, 13\penalty0 (2), 2024.
\newblock \doi{10.1186/s13677-023-00576-7}.
\newblock URL \url{https://doi.org/10.1186/s13677-023-00576-7}.

\bibitem[Irribarra et~al.(2024)Irribarra, Michell, Bermeo, and Kristjanpoller]{4}
Nicolás Irribarra, Kevin Michell, Cristhian Bermeo, and Werner Kristjanpoller.
\newblock A multi-head attention neural network with non-linear correlation approach for time series causal discovery.
\newblock \emph{Applied Soft Computing}, 165\penalty0 (C):\penalty0 107314, 2024.
\newblock \doi{10.1016/j.asoc.2024.112062}.
\newblock URL \url{https://doi.org/10.1016/j.asoc.2024.112062}.

\bibitem[Liu et~al.(2025)Liu, Liu, Chai, and et~al.]{8}
F.~Liu, S.~Liu, Y.~Chai, and et~al.
\newblock Enhanced mamba model with multi-head attention mechanism and learnable scaling parameters for remaining useful life prediction.
\newblock \emph{Scientific Reports}, 15:\penalty0 7178, 2025.
\newblock \doi{10.1038/s41598-025-91815-1}.
\newblock URL \url{https://doi.org/10.1038/s41598-025-91815-1}.

\bibitem[Malashin et~al.(2024)Malashin, Tynchenko, Gantimurov, Nelyub, and Borodulin]{7}
Ivan Malashin, Vadim Tynchenko, Andrei Gantimurov, Vladimir Nelyub, and Aleksei Borodulin.
\newblock Applications of long short-term memory (lstm) networks in polymeric sciences: A review.
\newblock \emph{Polymers}, 16\penalty0 (18):\penalty0 2607, 2024.
\newblock \doi{10.3390/polym16182607}.
\newblock URL \url{https://doi.org/10.3390/polym16182607}.

\bibitem[Mishra et~al.(2022)Mishra, Sriharsha, and Zhong]{6}
Abhinav Mishra, Ram Sriharsha, and Sichen Zhong.
\newblock Onlinestl: Scaling time series decomposition by 100x.
\newblock \emph{arXiv preprint}, 2022.
\newblock URL \url{https://arxiv.org/abs/2107.09110}.

\bibitem[Wang et~al.(2023)Wang, Shao, and Jumahong]{3}
W.~Wang, J.~Shao, and H.~Jumahong.
\newblock Fuzzy inference-based lstm for long-term time series prediction.
\newblock \emph{Scientific Reports}, 13:\penalty0 20359, 2023.
\newblock \doi{10.1038/s41598-023-47812-3}.
\newblock URL \url{https://doi.org/10.1038/s41598-023-47812-3}.

\bibitem[{World Health Organization}(2023)]{11}
{World Health Organization}.
\newblock World health statistics 2023: monitoring health for the sdgs, sustainable development goals, 2023.
\newblock URL \url{https://www.who.int/publications/i/item/9789240074323}.
\newblock [EB/OL].

\bibitem[Xiao et~al.(2023)Xiao, Liu, and Li]{9}
Degui Xiao, Jiahui Liu, and Jiazhi Li.
\newblock Falnet: flow-based attention lightweight network for human pose estimation.
\newblock \emph{Journal of Electronic Imaging}, 32\penalty0 (5):\penalty0 053008--053008, 2023.

\bibitem[Zhang et~al.(2020)Zhang, Lam, Li, and Han]{10}
Qi~Zhang, Jacqueline~CK Lam, Victor~OK Li, and Yang Han.
\newblock Deep-air: A hybrid cnn-lstm framework forfine-grained air pollution forecast, 2020.
\newblock URL \url{https://arxiv.org/abs/2001.11957}.

\end{thebibliography}

\end{document}